\documentclass{article}
\usepackage{spconf,amsmath,amssymb,graphicx,booktabs}
\usepackage{url}
\usepackage[english]{babel}
\title{A Discriminative Condition-Aware Backend for Speaker Verification}

\name{Luciana Ferrer$^1$, Mitchell McLaren$^2$\vspace{-0.3cm}\thanks{The research by authors at SRI International was funded through a development
contract with Sandia National Laboratories (SNL) (Subcontract\#
1758993/ DO 1872160). The views herein are those of the authors and do
not necessarily represent the views of the funding agencies.}}
\address{
  $^1$Instituto de Investigaci\'on en Ciencias de la Computaci\'on (ICC), CONICET-UBA, Argentina\\
  $^2$Speech Technology and Research Lab (StarLab), SRI International, USA}

\DeclareMathOperator{\Norm}{Norm}
\DeclareMathOperator{\softmax}{softmax}

\begin{document}
\ninept

\setlength{\abovedisplayskip}{6pt}
\setlength{\belowdisplayskip}{6pt}

\maketitle

\begin{abstract}
\vspace{-0.02cm}
We present a scoring approach for speaker verification that mimics the standard PLDA-based backend process used in most current speaker verification systems. However, unlike the standard backends, all parameters of the model are jointly trained to optimize the binary cross-entropy for the speaker verification task. We further integrate the calibration stage inside the model, making the parameters of this stage depend on metadata vectors that represent the conditions of the signals. We show that the proposed backend has excellent out-of-the-box calibration performance on most of our test sets, making it an ideal approach for cases in which the test conditions are not known and development data is not available for training a domain-specific calibration model. 
\end{abstract}
\vspace{-0.06cm}
\vspace{-0.03cm}
\section{Introduction}
\label{sec:intro}
\vspace{-0.16cm}
Most current speaker verification systems are composed of several separate stages. First, frame-level features that represent the short-time contents of the signal are extracted. These features are input to a deep neural network which is trained to optimize speaker classification performance on the training dataset. A hidden layer within that DNN is then used as a signal-level feature extractor. These new features, termed speaker embeddings or `x-vectors'~\cite{snyder2016deep}, are transformed using linear discriminant analysis (LDA), and then mean and length normalized. Next, probabilistic linear discriminant analysis (PLDA) is used to obtain scores for each speaker verification trial. Finally, a calibration stage is necessary to convert the scores produced by PLDA into proper log-likelihood ratios (LLRs) that can be thresholded or used directly to make decisions. This stage is usually trained to optimize a weighted binary cross-entropy objective which measures the overall quality of the scores as proper LLRs.

The procedure above has resulted in the best performing text-independent speaker verification systems to date on many different datasets. Nevertheless, this approach is likely to be suboptimal, since the only step that is trained to optimize actual speaker verification performance is the final stage of calibration. This problem has been widely acknowledged in the community and several papers have been published with different attempts to eliminate or integrate some of these stages. Angular and triplet losses \cite{li2018angular,huang:2018,bhattacharya2018,Novoselov2018is} have been used to train the speaker embedding extractor DNN instead of the standard cross-entropy with the goal of making the backend stages unnecessary. The cosine distance between embeddings resulting from these losses can be directly used to generate a score for each trial without the need for a separate backend. Nevertheless, since these losses do not directly measure speaker verification performance they do not result in calibrated scores, so a final calibration stage is still needed.

Recent papers \cite{snyder2016deep,Rohdin2018} proposed to use the binary cross-entropy as loss function during DNN training for end-to-end speaker verification. In \cite{snyder2016deep}, a DNN is used to obtain embeddings for each signal. The score for a trial is then computed using a simple function of the two embeddings with the same form as the PLDA scores, an idea that was first proposed in \cite{burget:icassp11}. The parameters for the embedding extractor and the scorer are trained jointly to optimize binary cross-entropy. In \cite{Rohdin2018}, the authors propose to use an architecture that mimics the previous i-vector \cite{Dehak11} pipeline for speaker verification, pretraining all its parameters separately and then fine tuning the full model to minimize binary cross-entropy. While these two approaches have the potential to result in well-calibrated scores, neither of the two papers show overall system performance, only discrimination performance.

In this paper we propose a backend approach that uses, as the two works mentioned above, the binary cross-entropy as the objective and a functional form that mimics that of the standard PLDA-based backend. In this initial work we do not aim for an end-to-end system. Instead, we take standard speaker embeddings as input. Our goal is to design a backend that can result in well-calibrated LLRs across a wide variety of conditions, without the need for further calibration. We show that the PLDA-like form, while leading to good discrimination performance, does not result in well calibrated scores across conditions, even for conditions similar to those included during training. We propose a modification to the functional form of the backend to take into account the conditions of the signals and show that this model results in a well-calibrated system across a wide range of conditions.

\vspace{-0.1cm}
\section{Standard PLDA-based Backend}
\vspace{-0.2cm}
Most state of the art speaker verification systems consist of an embedding extraction stage followed by a PLDA-based backend. The PLDA-based backend is in itself composed of several stages. First, linear discriminant analysis is applied to reduce the dimension of the embeddings while emphasizing speaker information and reducing other irrelevant information. Then, the global mean is subtracted and the resulting vectors are length normalized. Finally, PLDA is used to compute a score for each trial. While the training procedure for PLDA is somewhat involved and requires the use of an expectation-maximization algorithm, once parameters have been trained, scoring is done with a simple function of the two embeddings involved in the trial (see \cite{cumani2013pairwise} for a derivation). 

To summarize, the set of equations required to go from two individual embeddings, $x_1$ and $x_2$, to a score $s$ for the trial are:
\begin{eqnarray}
\tilde x_i & = & \Norm( Px_i+\mu), \forall i \in \{1,2\} \label{eq:lda} \\
s & = & 2 \tilde x_1 ^T \Lambda \tilde x_2 + \tilde x_1^T \Gamma \tilde x_1 + \tilde x_2^T \Gamma \tilde x_2 + (\tilde x_1 + \tilde x_2)^T c + k \label{eq:plda_score}
\end{eqnarray}
where $P$ is the LDA projection matrix, $\mu$ is the global mean after LDA projection, $\Norm$ performs length normalization, and $\Lambda$, $\Gamma$, $c$ and $k$ are derived from the parameters of the PLDA model using Equations (14) and (16) in \cite{cumani2013pairwise}. 

The final step in the backend is calibration. This is the only step that is trained to optimize actual speaker verification performance, including calibration. PLDA scores are computed as the logarithm of the ratio between the likelihood for the hypothesis that the speakers in the two signals in the trial are the same and the likelihood of the hypothesis that the speakers are different. That is, the score is defined as a log-likelihood ratio (LLR). Yet, in practice, the scores produced by PLDA are far from being proper LLRs. This is due to the fact that PLDA's assumption do not exactly hold in practice. Hence, the LLRs produced by the model are not well-calibrated or, in other words, they are not proper LLRs, and a final stage of calibration is required. 
It is possible to use the raw scores from PLDA to make speaker verification decisions by tuning a threshold on some development data for the specific application of interest. Yet, in many cases, like in forensic applications, 
when the operating point is not defined a priori, it is necessary for the system to output proper LLRs which can then be thresholded using Bayes rule for any cost of interest or directly used as  stand-alone interpretable values.

The standard procedure for calibration in speaker verification is to use linear logistic regression, which applies an affine transformation to the scores, training the parameters to minimize binary cross-entropy \cite{brummer2013likelihood}.  The objective function to be minimized is given by
\begin{equation}
C_\pi = -\frac{\pi}{T} \sum_{k \in \cal T}  \log(q_k) - \frac{1-\pi}{N} \sum_{k \in \cal N}  \log(1-q_k), \label{eq:crossent}
\end{equation}
where
\vspace{-0.61cm}
\begin{eqnarray}
q_k & = &\sigma\left(l_k + \log (\pi /(1-\pi))\right), \\
l_k & = & \alpha s_k + \beta \label{eq:cal}
\end{eqnarray}
where $s_k$ is the score for  trial $k$ given by Equation (\ref{eq:plda_score}), $\sigma$ is the sigmoid function, and $\alpha$ and $\beta$ are the calibration parameters, trained to minimize the quantity in Equation (\ref{eq:crossent}).

To summarize, Equations (\ref{eq:lda}), (\ref{eq:plda_score}) and (\ref{eq:cal}) show the pipeline that is applied to the embeddings in the standard PLDA-based backend. The parameters involved in these equations are all trained separately, freezing the parameters of the previous steps in order to obtain input data to train the next step. 

\vspace{-0.1cm}
\section{Proposed Discriminative Backend}
\vspace{-0.2cm}
We propose a backend with the same functional form as the PLDA-backend explained in the previous section, but where all parameters are optimized jointly, in a manner similar to the one used in \cite{Rohdin2018} (though, note that in this paper we only optimize jointly up to the backend stage instead of the full pipeline, as in Rohdin's paper). We first initialize all parameters in Equations (\ref{eq:lda}), (\ref{eq:plda_score}) and (\ref{eq:cal}) as in the standard PLDA-based backend. Then, we fine tune the parameters to optimize the cross-entropy in Equation (\ref{eq:crossent}) using some variant of stochastic gradient descent. To this end, we need to define mini-batches that contain both negative and positive samples. This is done by randomly selecting N speakers for each mini-batch. Then, two random samples from each of those speakers are chosen. All possible trials between the 2N selected samples are used to compute the cross-entropy, after excluding all same-session target trials and different-domain impostor trials. We found that these two restrictions were important to get good calibration performance. We refer to this as the Discriminative PLDA (DPLDA) backend.

As we will see in the results section, the approach above leads to good discrimination performance over a large set of conditions, sometimes improving over the baseline, in agreement with results found in \cite{Rohdin2018}. Nevertheless, calibration performance of this approach is far from optimal on many conditions. This is the same phenomenon observed for PLDA, which generally has reasonable discrimination performance on unseen domains (though usually suboptimal compared to that of a system adapted to the specific domain) but extremely bad overall calibration performance. This problem is usually fixed by training a specific calibration model for each domain of interest, which requires having at least some domain-specific labeled data. 

In this work, we assume that no domain-specific data is available for system adaptation or for training a calibration model. This also means that a domain-specific decision threshold cannot be learned. Hence, we aim to design the best possible out-of-the-box system for unknown conditions for which the  score produced can be thresholded using the theoretically optimal threshold assuming the scores are proper LLRs (see, for example, Equation (6) in \cite{van2007introduction}). In order to achieve this goal, the calibration parameters have to depend on the signal's conditions. We propose that this dependence be achieved by having the calibration scale and shift, $\alpha$ and $\beta$ in Equation (\ref{eq:cal}), be functions of metadata vectors, $z_1$ and $z_2$, for each of the signals in a trial:
\begin{eqnarray}
\alpha  =  2 z_1^T \Lambda_\alpha z_2 + z_1^T \Gamma_\alpha z_1+ z_2^T \Gamma_\alpha z_2 + (z_1 + z_2)^T c_\alpha + k_\alpha \label{eq:alpha} \\
\beta   =  2 z_1^T \Lambda_\beta z_2 + z_1^T \Gamma_\beta z_1 + z_2^T \Gamma_\beta z_2 + (z_1 + z_2)^T c_\beta + k_\beta \label{eq:beta} 
\end{eqnarray}
In our implementation, all parameters in these equations are initialized to 0 except the $k$ values that are initialized with the  global calibration parameters trained using linear logistic regression.

The key component of this model are the metadata vectors $z_i$. Ideally, they should be a general representation of the signal's conditions (channel, language, background noise, reverberation, etc) except for the identity of the speaker. That is, we want $\alpha$ and $\beta$ to be condition-dependent but speaker-independent, leaving the score $s_k$ to contain the information about the trial's class (same-speaker versus different-speaker). To this end, we propose to use a separate DNN trained to predict the signal's condition. The pre-activations in a bottleneck layer from that DNN, which we will call $m_i$, are then further transformed with a trainable linear transformation followed by a softmax. The logarithm of the result is used as metadata vector: 
\begin{equation}
z_i = \log \softmax(W m_i) \label{eq:meta}
\end{equation}
Several other options were tested to transform $m_i$ into $z_i$: adding a bias terms, using length-normalization, no transformation, softmax without the logarithm and relu. None of these alternatives proved to be better in our experiments than the logarithm of the softmax transformation. Also, the trainable transformation proved to be essential to obtain good performance. Using the pre-activations (or activations) from the condition DNN directly as $z$ values with or without log softmax led to suboptimal results. In our experiments the $W$ parameter is trained jointly with the rest of the model. This is the only parameter that is initialized randomly using a normal distribution centered at 0.0 with standard deviation of 0.5.

The idea of using metadata to condition the calibration model has been explored in a few works before. In most cases, the metadata was assumed to be discrete and known (or estimated separately) during testing \cite{FerrerEtAl:Eurospeech2005,Solewicz:Eurospeech2005,solewicz:ASLP07,FerrerEtAl:icassp2008}. The calibration parameters are then conditioned on the discrete metadata values. The Focal Bilinear toolkit \cite{FocalBilinear} implements a version of metadata-dependent calibration where the calibrated score is a bilinear function of the scores and the metadata vector, which is assumed to be composed of numbers between 0 and 1. More recently, we proposed an approach called trial-based calibration (TBC) where calibration parameters are trained independently for each trial using a subset of the development data \cite{mclaren2014trial,Ferrer:aslp18} selected using a model trained to estimate the similarity between the conditions of two samples. This approach, while successful, is quite computationally expensive and requires tuning a few different parameters in order to obtain good performance. In our proposed model both discrete (in the form of one-hot vectors) and continuous metadata can be used as the $z$ vectors and the functional form is a generalization of all previous approaches, except for TBC. In addition, in our proposed approach, the calibration model is trained jointly with the rest of the backend parameters, while in all previous approaches the calibration step was trained separately.

\vspace{-0.1cm}
\section{Experimental Setup}
\vspace{-0.2cm}
In this section we describe the system configuration and datasets used for our experiments. 

\vspace{-0.3cm}
\subsection{Speaker Recognition System}
\label{sec:system}
\vspace{-0.2cm}
The proposed backend uses standard x-vectors as input \cite{KaldiRecipe17}. 
The input features for the embedding extraction network are power-normalized cepstral coefficients (PNCC)~\cite{kim:12} which, in our experiments, gave better results than the more standard mel frequency cepstral coefficients (MFCCs). We extract 30 PNCCs with a bandwith going from 100 to 7600 Hz and root compression of 1/15. The features are mean and variance normalized over a rolling window of 3 seconds. Silence frames are discarded using a DNN-based speech activity detection system.

System training data included 234K signals from 14,630 speakers. This data was compiled from NIST SRE 2004--2008, NIST SRE 2012, Mixer6, Voxceleb1, and Voxceleb2 (train set) data. Voxceleb1 data had 60 speakers removed that overlapped with Speakers in the Wild (SITW). All waveforms were up- or down-sampled to 16 KHz before further processing.  In addition, we down-sampled any data originally of 16 kHz or higher sampling rate (74K files) to 8 kHz before up-sampling back to 16 kHz, keeping two ``raw'' versions of each of these waveforms. This procedure allowed the embeddings system to operate well in both 8kHz and 16kHz bandwidths.

Augmentation of data was applied using four categories of degradations as in~\cite{mclaren:odyssey18}, including music and noise, both at 10 to 25 dB signal-to-noise ratio, compression, and low levels of reverb. We used 412 noises compiled from both freesound.org and the MUSAN corpus. Music degradations were sourced from 645 files from MUSAN and 99 instrumental pieces purchased from Amazon music. For reverberation, examples were collected from 47 real impulse responses available on echothief.com and 400 low-level reverb signals sourced from MUSAN. Compression was applied using 32 different codec-bitrate combinations with open source tools. We augmented the raw training data to produce 2 copies per file per degradation type (randomly selecting the specific degradation and SNR level, when appropriate) such that the data available for training was 9-fold the amount of raw samples. In total, this resulted in 2,778K files for training the speaker embedding DNNs. 

The architecture of our embeddings extractor DNN follows the Kaldi recipe~\cite{KaldiRecipe17}. The DNN is implemented in Tensorflow, trained using an Adam optimizer using chunks of speech between 200 and 350 milliseconds. Overall, we extract about 4K chunks of speech from each of the speakers. 
DNNs were trained over 4 epochs over the data using a mini batch size of 96 examples. 
We used dropout with a probability linearly increasing from 0.0 up to 0.1 at 1.5 epochs then linearly decreasing back to 0.0 at the final iteration. The learning rate started at 0.0005, increasing linearly  after 0.3 epochs reaching 0.03 at the final iteration while training simultaneously using 8 GPUs.

The training data for the PLDA or DPLDA backends was a subset of the training data used for the speaker embeddings DNN including a random half of the speakers (for expedience of experimentation) and excluding all signals for which no information about the recording session could be obtained and all speakers for which a single session was available.  In this case, we use full segments to train the backend rather than chunks and SNR level of 5dB for augmentation (using this SNR on the PLDA backend led to marginally better results than using 10-25dB SNR, as for the embeddings extractor). Both for PLDA and DPLDA, the LDA dimension is set to 200. 

Beside the training data above, we add two datasets for backend training, FVCAus and RATS.  FVCAus is composed of interviews and conversational excerpts from over 500 Australian English speakers from the forensic voice comparison dataset \cite{morrison2015forensic}. Audio was recorded using close talking microphones. RATS is composed of telephone calls in five non-English languages from over 300 speakers. We only used the source data (not retransmitted) of the DARPA RATS program \cite{ref:rats_set} for the SID task. 


The condition DNN used to generate the embeddings $m_i$ which are used as input to obtain the metadata vectors $z_i$ (Equation \ref{eq:meta}) has two layers of 100 and 10 nodes with relu activations and batch normalization. The classes used at the output layer are given by the domain (Voxceleb, Mixer, Switchboard, FVCAus or RATS) concatenated with the degradation type and, when available, any further information about the condition of the signal (channel type, language, and speech style). Note that the classes are then extremely different in terms of granularity. All Voxceleb data is grouped into one class per degradation type, while Mixer data has much finer grained labels. While this is clearly suboptimal, it seems to work well in our experiments. An alternative we are currently pursuing is to obtain the embeddings using unsupervised techniques.

\vspace{-0.3cm}
\subsection{Two-stage training}
\label{sec:two-stage}
\vspace{-0.2cm}
Our backend training data, described in Section \ref{sec:system}, is highly imbalanced: 53\% comes from voxceleb collections, 25\% from SRE and Mixer collections, 11\% from Switchboard, 6\% from RATS, and 4\% from FVCAus. This causes a problem when learning the calibration part of the model, since parameters cannot be robustly learned for the underrepresented conditions. For this reason, we implement a two-stage training procedure. We use all the training data for the first few iterations, then freeze the parameters up the score generation stage (Equation \ref{eq:plda_score}), subset the training list to use a balanced set of samples with similar representation for all five domains and continue training the calibration parameters (Equations \ref{eq:alpha}, \ref{eq:beta} and \ref{eq:meta}). This allows the model to focus on improving metadata-dependent calibration once the discriminative part of the model has converged. 

\vspace{-0.3cm}
\subsection{Datasets}
\label{sec:evaldata}
\vspace{-0.2cm}

We use several different datasets for development and evaluation of the proposed approach. Table \ref{tab:dsets} shows the statistics for all sets.
The {\bf SITW} dataset contains speech samples in English from open-source media~\cite{sitwdb} including naturally occurring noises; reverberation; codec; and channel variability. 
The {\bf SRE16} dataset \cite{sre16} includes variability due to domain/channel and language mismatches. 
We use the CMN2 subset of the {\bf SRE18} dataset \cite{sre18}, which has similar characteristics to the SRE16 dataset, with the exception of focusing on different languages, and including speech recorded over VOIP instead of just PSTN calls.
The {\bf LASRS} corpus is composed of 100 bilingual speakers from each of three languages, Arabic, Korean and Spanish \cite{Beck2004}. Each speaker is recorded in two separate sessions speaking English and their native language using several recording devices.  
Finally, the {\bf FVCCMN} is composed of interviews and conversational excerpts from over 68 female Chinese speakers from the forensic voice comparison dataset \cite{zhang2011forensic}, which were cut to durations between 10 and 60 seconds. Recordings were made with high-quality lapel microphones.

\begin{table}[!t]
\footnotesize
\centering
\vspace{-0.15cm}
\caption{The development and evaluation datasets with number of speakers and target/impostor (tgt/imp) trial counts.}
\vspace{0.1cm}
\begin{tabular}{l|ccc|ccc}
	Dataset & \multicolumn{3}{c|}{Dev Split} &  \multicolumn{3}{c}{Eval Split} \\
	              & Speakers & \#tgt & \#imp & Speakers & \#tgt & \#imp  \\ 
	\midrule
	SITW  &  119 & 2.6k & 335k &  180 & 3.7k &  717k \\
	SRE16  &  20 & 3.3k & 13.2k &  201 & 27.8k & 1.4m \\
	SRE18  &  35 & 6.3k & 80.2k  &  289 & 48.5k & 1.6m \\
	FVCCMN & - & - & - & 68 & 16.4K & 1.1m \\
	LASRS & - & - & - &  333 & 41k & 4.8m \\
	\bottomrule
\end{tabular}
\label{tab:dsets}
\end{table}

%
%
%
%

SITW, SRE16 and SRE18 have a well-defined development sets. We use those 3 sets to tune the parameters of our models. The rest of the sets are used for evaluation of the final systems. For SITW, SRE16 and SRE18 we use the 1-side enrollment trials defined with the datasets. For LASRS and FVCCMN we create exhaustive trials excluding same-session trials.

\vspace{-0.1cm}
\section{Results}
\vspace{-0.2cm}
We show results in terms of Cllr. This metric \cite{Brummer:csl06} measures the quality of the scores as LLRs using a logarithmic cost function and is affected both by the discrimination and calibration performance of the system. A very discriminant system can have a high Cllr if the calibration is wrong (ie, if the scores do not represent proper LLRs for the task). Such a system would lead to bad decisions when thresholded with the theoretically optimal threshold for the cost function of interest. A way to generate good decisions for such a system would be to obtain representative development data and either calibrate the system with it  or, if the application costs are known before hand,  empirically find the optimum threshold for those costs. 

In this work, we aim to obtain a system that results in well-calibration scores across a large variety of conditions without requiring explicit development data for each case. To measure whether we are succeeding in this goal, we need to separate the effect of the discrimination and the calibration performance of the system. This is done by obtaining the minimum Cllr that can be achieved with the system's scores for a certain test set using a monotonic transformation obtained using the pool adjacent violators  (PAV) algorithm \cite{brummer:pav}. The difference between the actual Cllr and minimum Cllr for the system indicates the effect of miss-calibration. If the two values are equal, then the system is perfectly calibrated and the scores produced by it are proper LLRs.

\begin{figure}[!t]
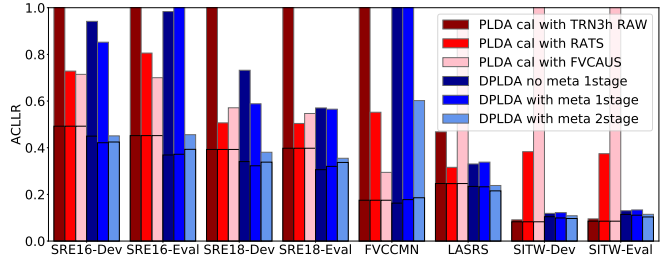

\centering
\hspace{-0.3cm}
\includegraphics[width=1.02\columnwidth]{{{figs/TABLE.plda_vs_dplda.ACLLR}}}
\vspace{-0.5cm}\caption{Actual CLLR (bar height) and minimum Cllr (black line inside bars) for different PLDA and DPLDA systems on all test sets. Bars taller than 1.0 are cut to allow better resolution of all other results. } 
\label{fig:res}
\end{figure}

Figure \ref{fig:res} shows the actual and minimum Cllr values for development and evaluation dataset for different systems. All development decisions (dimensions, training hyperparameters, etc) were made based on the average performance in all three development sets plus FVCAus and RATS (even though performance in those sets is optimistic, since they were seen during training). Only the decision to add FVCAus and RATS to the training list was made considering the evaluation datasets, since we realized that some conditions were not represented in the training data, which restricted the robustness of the system. During development we concluded that including the $\Gamma_\alpha$ and $\Gamma_\beta$ terms in Equations (\ref{eq:alpha}) and (\ref{eq:beta}) did not improve results. Hence, those terms are not used in our experiments. The results shown correspond to the best of 5 models  run with different seeds for the chosen architecture, selected based on the development sets' performance. The seed seems to have a significant effect in the calibration performance of the final model (not in discrimination). This is a weakness of the approach that we plan to work on in the future.

For the three PLDA systems the LDA and PLDA parameters are trained using only the training data, without adding FVCAus and RATS since adding those sets slightly degrades discrimination performance of this system on our development sets. We show three options for training the calibration stage for the PLDA system: using TRN3h RAW which consists of 300 speakers from the raw part of the training set (using more speakers does not help and including the degraded part hurts performance), using only RATS data and using only FVCAus data. Note that no calibration set is optimal for all test sets. Merging the three sets leads to a trade-off in performance which highly depends on the proportion of each dataset used (results not shown). Note that the discrimination performance of the baseline system is not affected by the calibration model, since this model is a single monotonic transformation for each test set.

Finally, we show three DPLDA systems. The first one is trained in one stage without including metadata ($\alpha$ and $\beta$ are scalars). The second system introduces the use of metadata but it is also trained in one stage. The third system is like the second one but it is trained with the two-stage algorithm explained in Section \ref{sec:two-stage}. For the last two systems, the $W$ matrix reduces the condition embeddings from 10 to 5 dimensions. In all three cases the number of total iterations is decided based on development results. We can see that the best DPLDA approach is to use metadata with 2-stage training. This method provides the best or close to the best calibration on all test conditions except one, FVCCMN. This dataset is quite different from all our training data. While it is similar in terms of acoustic conditions to FVCAus, it consists only of Chinese speech. While around 1\% of the training samples are in Chinese (all from SRE04 and SRE06 datasets), they are all recorded over a telephone channel, not in the extremely clean acoustic conditions of the FVCCMN dataset. We believe this is a case where the system should be able to reject the trials for being severely mismatched to the training data. We plan to pursue this research direction in the near future. 

\vspace{-0.1cm}
\section{Conclusions}
\vspace{-0.2cm}
We presented a novel approach for speaker verification scoring using speaker embeddings. The approach consists on a series of operations that mimic the standard PLDA-backend followed by calibration. The parameters of the model are learned jointly to optimize the overall speaker verification performance of the system, directly targeting the loss of interest in the speaker verification task. Further, we propose a formulation for the use of metadata describing the conditions in the signal to condition the calibration parameters. We show that the proposed approach significantly improves performance  over the standard PLDA backend on a wide variety of test conditions, leading to a robust backend that does not require specific development data for calibration. 


\bibliographystyle{IEEEbib}
\bibliography{./all-short}

\end{document}